\documentclass{article}

\usepackage{arxiv}

\usepackage[utf8]{inputenc} % allow utf-8 input
\usepackage[T1]{fontenc}    % use 8-bit T1 fonts
\usepackage{hyperref}       % hyperlinks
\usepackage{url}            % simple URL typesetting
\usepackage{booktabs}       % professional-quality tables
\usepackage{amsfonts}       % blackboard math symbols
\usepackage{nicefrac}       % compact symbols for 1/2, etc.
\usepackage{microtype}      % microtypography
\usepackage{lipsum}
\usepackage{graphicx}

\title{Meta-learning of textual representations\thanks{This work was partially supported by CONACyT under grant \emph{Aprender Objetos de Internet para Buscarlos con un Robot}, ID: CB-250938.}}

\author{
Jorge G. Madrid\textsuperscript{1} \quad Hugo Jair Escalante\textsuperscript{1,2,3} \quad Eduardo Morales\textsuperscript{1}\\
 \textsuperscript{1}INAOE, Mexico\\
 \textsuperscript{2}CINVESTAV, Mexico\\
 \textsuperscript{3}ChaLearn, USA\\
 \{jgmadrid, hugojair, emorales\}@inaoep.mx
}

\begin{document}
\maketitle

\begin{abstract}
Recent progress in AutoML has lead to state-of-the-art methods (e.g., AutoSKLearn) that can be readily used by non-experts to approach \emph{any} supervised learning problem. Whereas these methods are quite effective, they are still limited in the sense that they work for tabular (matrix formatted) data only. 
%are already available, nonetheless none of them concentrate on the challenges of Natural Language Processing. 
This paper describes one step forward in trying to automate the design of supervised learning methods in the context of text mining. %We approach the problem of automatically obtaining a representation for text mining tasks starting from raw text. 
We introduce a meta learning methodology for automatically obtaining a representation for text mining tasks starting from raw text.  We report experiments considering 60 different textual representations and more than 80 text mining datasets associated to a wide variety of tasks. Experimental results show the proposed methodology is a promising solution to obtain highly effective off the shell text classification pipelines. 

\end{abstract}

% keywords can be removed

\section{Introduction}
Nowadays, the success of machine learning systems relies on the knowledge of human-experts that according to their experience design and test multiple models extensively to select the best modeling option.  Although effective, %However,
this strategy is not only time consuming but also impractical since  an expert is not always available. % in many domains, 
This has motivated the  increasing demand for easy-to-use automated machine learning solutions solutions. In this context, AutoML is the field of research aiming to generate machine learning models without any human supervision. 
%aims to provide these solutions by taking the human out of the loop.
Recent progress in AutoML has lead to quite effective and competitive solutions when dealing with tabular (matrix formatted) data, see e.g.,~\cite{Escalante:2009:PSM:1577069.1577084,autoweka,feurer2015efficient}. However, these techniques are still limited in the sense that they require the user to transform raw data into a tabular representation. This step, relying heavily on the expertise of users. 

Text classification is one of the most studied tasks in Natural Language Processing (NLP), this is because of the number of applications that can be approached as text classification problems (e.g. sentiment analysis, topic labeling, spam detection, and author profiling among others). Many techniques for pre-processing, feature extraction, feature selection and document representation have been developed over the last decades. Despite all this progress by the NLP community,  it is still an expert who designs the pipeline for text classification systems that includes one or many of such techniques, each of which  usually requiring the fine-tuning of hyperparameters.

In this work we take a step towards the automated generation of the classification pipelines for text classification by focusing on the representation. Thus, our goal is to automate the process of determining the best representation to approach a text classification task starting from raw text. % This is relevant for a number of reasons, including the fact 
Unlike other data type, language/text provides an unstructured and rich source of information, hence selecting the adequate representation for text will have a direct impact in the performance of text mining solutions. In fact, representing texts has been one of the most studied venues in NLP. We propose a meta-learning solution to automatically determine the best representation given a dataset of raw text. This is a first step towards the full automation of the generation of text classification pipelines. We propose a number of meta features some of which are extracted directly from raw text, and most of them not used before for meta-learning. We report experimental results considering 60 textual representations and more than 80 text mining tasks. %To the bes
%, which features to extract is one of the most studied questions in NLP, the selection of an adequate representation is fundamental for successfully solving a classification task and usually has a greater impact than other parts of the pipeline such as the classification model. 
Our results show that the proposed meta features successfully characterize text mining tasks and that an AutoML solution for text mining (AutoText) is feasible. Our work is among the first to approach text classification via meta-learning from raw data and it is by far the largest study on meta learning in the context of text mining. %We perform experiments of larger scale than previous work (in 81 publicly available corpora) and include more meta-data by extracting information not only from pipelines but also from the \textit{raw} text.

\section{Related Work}
In the context of text mining, few works have explored the automated selection of different parts of classification pipelines. With experiments in the Reuters-21578 corpus, Lam \& Lai\cite{lam2001meta} proposed to characterize documents with 9 (meta) features and to predict the classification error of different models using data from a previous phase, thus recommending a classification model. More recently,  \cite{yogatama2015bayesian} searched for text representations with Bayesian Representation \cite{snoek2012practical}.
%\todo[author=HJ,inline]{include reference for BO}
Their search space was limited to only word n-grams and experiments were performed in 8 datasets: 4 sentiment analysis tasks and 4 topic classification tasks. Nevertheless, they outperformed every linear classifier reported until their publication date. Despite their limited scope, given the lack of data and computational resources of the time, this work represents one of the first meta-learning approaches for text classification.
%\todo[author=HJ,inline]{Servir\'ia mucho poner los n\'umeros de features y tareas usados en trabajo preliminar}

Other works have explored different meta-learning approaches for text classification in small-scale, for example, Gomez, et. al., \cite{gomez2017evolutionary} addressed the problem with evolutionary computation methods and using 11 meta-features. In a broader approach Ferreira \& Brazdil\cite{ferreira2018workflow} recommend \textit{pipelines} with Active Testing Method, in their work they also present statistical analysis of 48 preprocessing methods and 8 classifiers. 

Another related work is that by \cite{pinto2008clustering} where a set of features derived from the text was proposed for characterizing short-text corpora  in the context of clustering. Although the goal of such reference was to characterize the harness of text corpora and not AutoML, such work is relevant because their features inspired some of the meta-features considered in this paper. %as they used a sort of meta-features.  by how difficult is it to classify its text documents. While such set of features wasn't design for AutoML purposes it characterizes to some degree certain aspects of written language. 
In Section 3 we describe how we combined this set of features with other proposed ones for characterizing text collections in the context of automatic text mining.  %from more traditional meta-learning methods and others from NLP techniques.

%On the other hand, 
Meta-learning has been studied for a while in the broad machine learning context~\cite{bozdogan1987model,bengio2000gradient,bergstra2011algorithms,vilalta2002perspective,vanschoren_meta_2018}. However,  it is only recently that it has become a mainstream topic, this mainly because of its  successes in several tasks. For instance, Feurer, et. al., \cite{feurer2015initializing} successfully used a set of meta-features to warm-start a hyper-parameter optimization technique in the popular state-of-the-art  AutoML solution \textit{Autosklearn}.  %(this method has won  a series of   AutoML challenges~\cite{guyon2017analysis}). 
Likewise, the success of deep learning together with the difficulty in defining appropriate architectures and hyperparameters for users, has motivated a boom on neural architecture search, where  meta-learning is becoming common~\cite{NAS18}.

In this paper we propose a novel approach to meta-learning of text representations. We propose a novel set of meta-features, comprising standard meta-features from the machine learning literature, features that have been used for other problems than meta-learning and novel meta-features that have not been used previously.
some of which are derived directly from raw text and aim at capturing complex language patterns. We approach the problem of recommending textual representations. Whereas this problem has been addressed in previous work, such references have considered only a few representations and a very limited number of meta-features (up to 11).

To the best of our knowledge this is the largest scale study on meta-learning in the context of text mining. Whereas results are promising, please note that this is only a first step towards the ultimate goal of automating the text mining process. 
%\todo[author=HJ,inline]{verificar n\'umeros de este p\'arrafo}
%Compared to previous work on meta-learning in the context of text mining, our approach comprises novel meta-features that characterize.  
%Similarly, for this work it is expected that our proposed set of meta-features from text corpora will provide an accurate description of them which could be used to perform an analogous process for finding a text representation in a discretized search space.

\section{Recommending textual representations}
%\todo[author=HJ,inline]{cambiar a t\'itulo m\'as descriptivo del trabajo}
We introduce a meta-learning method that takes as input the labeled raw text from a corpus associated to a text classification task and automatically selects a representation.
The method recommends vector representations for text classification tasks based on which one worked best for \textit{similar} tasks.  In order to do so we define a set of meta-features and perform extensive experiments on 81 different text classification tasks. Although this approach is common within meta-learning~\cite{vanschoren_meta_2018},  it has not been widely explored for text classification. In fact, previous work (see Section 2) %related work, described in section 2, h
has considered small subsets of generic meta-features.

Table \ref{tab:rep} sums up the feature extraction methods that with some pre-processing processes or hyper-parameters gives a to a total of 60 representations, while not exhaustive, our work is the first to consider representations not only based on simple features, but also those based on topic modeling, embeddings, and semantic analysis. Furthermore, the output of our method can be useful for both human experts designing text classification pipelines and for complementing other optimization methods for AutoML (e.g. it can be used for warm-starting Bayesian Optimization\cite{feurer2015initializing} for a wider search space of text representations or easily combined with existing AutoML solutions).

\begin{table}[htbp]
\caption{Representations considered}
\begin{center}

\begin{tabular}{ll}
\hline
Features & Hyper-parameters \\
\hline
N-grams & [words, char], stop\_words[None, ‘English’], range[1,3], weight[bi,tf,tfidf] \\ 
LDA & stop\_words[None,’English’] \\ 
LSA & stop\_words[None,’English’], weight[tf,tfidf] \\ 
LIWC &  \\ 
W2V & pre\_trained[True,False], vector[mean, sum] \\ 
\hline
\end{tabular}
\end{center}

\label{tab:rep}
\end{table}

%\todo[author=HJ,inline]{Aqu\'i falta completar un p\'arrafo explicando la idea intuitiva, motivaci\'on y posible impacto. }
%based on results from previous experiments, then a fixed model is used for classifying unseen documents. 

The proposed method comprises 2 stages, an offline phase where it \textit{learns how to learn} and a predicting phase where it uses the data collected in phase 1 to recommend a text representation for classifying. 
%\todo[author=HJ,inline]{Podr\'ias incluir una figura?}

A human-expert uses knowledge acquired in the past when a new task is presented, equivalently, \textit{meta-learning} imitates this reasoning. Our method applies meta-learning to learn from the performance of different representations on a number of corpora. Namely, we defined 72 meta-features to characterize 81 text corpora and performed an exhaustive search for the performance of 60 representations. A \textit{knowledge base} is built associating the performance of each representation with a task, described by the vector of meta-features. Traditionally, meta-features extract meta-data from a dataset such as statistics of its distribution or simple characteristics like the number of classes and attributes, in our proposed set we contemplate this type of features as well as other attributes extracted directly from the raw text. The proposed meta-features are described below. For clarity we have divided them in groups. %We propose a set of 73 meta-features combining meta-learning  traditional features with NLP  ones. Below we organized them in groups and provide a brief description.
\begin{itemize}
    \item \textbf{General meta-features.} %We take 2 simple measures from the corpus, 
    The \textit{number of documents} and the \textit{number of categories}.
    \item \textbf{Corpus hardness.} %These aim to capture information on the \emph{hardness} of text corpora, they were originally used in \cite{pinto2008clustering} 
    Most of these originally used in \cite{pinto2008clustering} to determine  the hardness of short text-corpora. %,  and take into account different features from text.

        \emph{Domain broadness.} Measures related to the thematic broadness/narrowness of words in documents. 
        %Measures that capture the broadness degree of the corpus, a narrow broadness includes terms closely related to each other while a wide one has more diversity on its terms. For instance, we would expect a corpus of several news categories such as 20Newsgroups to have a wider broadness than a corpus of hotel reviews. 
        %In our proposed set of meta-features 
        We included measures based on the vocabulary length and overlap: \textit{Supervised Vocabulary Based (SVB)}, \textit{Unsupervised Vocabulary Based (UVD)} and \textit{Macro-averaged Relative Hardness (MRH)}.
        
        \emph{Class imbalance.} %Refers to the document distribution across the classes. A simple 
        \textit{Class Imbalance (CI)} ratio.
        
        \emph{Stylometry.} %The writing style of a text usually attributed to a specific author. 
        \textit{Stylometric Evaluation Measure (SEM)}
        
        \emph{Shortness.} %Features based on the length of the text or the vocabulary used in the corpus. 
        \textit{Vocabulary Length (VL)}, \textit{Vocabulary Document Ratio (VDR)} and  average \textit{word length}.
    \item \textbf{Statistical and information theoretic.} %Similarly to previous work 
    We derive meta-features from a document-term matrix representation of the corpus.
    %Analogously to related work we described  the corpus documents with a document-term matrix and calculated the following meta-features:
    
         \textit{min, max, average, standard deviation, skewness, kurtosis, ratio average-standard deviation, and entropy of:}  vocabulary distribution, documents-per-category and  words-per-document:
        
        \emph{Landmarking.} 70\% of the documents are used to train 4 simple classifiers and their performance on the remaining 30\% was used based on the intuition that  some aspects of the dataset can be inferred: \textit{data sparsity - 1NN, data separability - Decision Tree, linear separability - Linear Discriminant Analysis, feature independence Na\"ive Bayes}. The \textit{percentage of zeros} in the matrix was also added as a measure for sparsity.
        
        \emph{Principal Components (PC) statistics}. %Another frequent group of meta-features come  from  
        Statistics derived from a PC analysis: 
        %from the Principal Component Analysis of the data, in our representation the components are the most relevant terms for a corpus, we included the following measures: 
        \textit{pcac} from~\cite{gomez2017evolutionary}; for the first 100 components, the same statistics from documents per category and their \textit{singular values sum, explained ratio and explained variance}, and for the first component its \textit{explained variance}.
    \item \textbf{Lexical features.} %As part of our characterization 
    We incorporated the distribution of parts of speech tags. We intuitively believe that the frequency of some lexical items will be higher depending on the task associated to a corpus, for instance a corpus for sentiment analysis may have more adjectives while a news corpus may have less. We tagged the words in the document and computed the average number of \textit{adjectives, adpositions, adverbs, conjunctions, articles, nouns, numerals, particles, pronouns, verbs, punctuation marks} and \emph{untagged words}  in the corpus. 
    \item \textbf{Corpus readability.} Statistics from text that determine readability, complexity and grade from textstat library\footnote{https://github.com/shivam5992/textstat}: \textit{Flesch Reading Ease, SMOG grade, Flesch-Kincaid grade level, Coleman-Liau index, automated readability index, Dale-Chall readability score, the number of difficult words, Linsear Write formula, Fog scale,} and \emph{estimated school level to understand the text}.
\end{itemize}
Apart from general, statistical and PC based, the rest of the listed features have not been used in a meta-learning context. After the offline phase takes place, for a new task the same meta-features are extracted and compared with the prior knowledge, to recommend a representation. We considered 4 strategies that leverage learned experiences and make predictions for a new task, these are described below % in Table~\ref{tab:variantes}.
%\begin{table}[htbp]
%\caption{Strategies for recommending textual representations.}
%\begin{center}
%\footnotesize{
%\begin{tabular}{rc}
%\hline
%Id & Description \\ \hline
%(1) &  Using directly the representation with best performance of the \textit{nearest} corpus \\ 
%(2) & Predicting the representation as a classification problem, where each representation is a class \\
%(3) & Predicting the performance for every representation and selecting the one with the smallest error \\
%(4) & Predicting the rank of each representation and selecting the one with best predicted rank\\ \hline
%\end{tabular}}
%\end{center}
%\label{tab:variantes}
%\end{table}

\begin{itemize}
    \item \textbf{(1)} Using directly the representation with best performance of the \textit{nearest} corpus.
    \item \textbf{(2)} Predicting the representation as a classification problem, where each representation is a class.
    \item \textbf{(3)} Predicting the performance for every representation and selecting the one with the smallest error.
    \item \textbf{(4)} Predicting the rank of each representation and selecting the one with best predicted rank.
\end{itemize}{}

For strategies 2-4 different classification and regression models were tested,  a Random Forest classifier was selected for strategy 2 and Random Forest regressor for both strategies 3 and 4. Once the representation is chosen, an SVM classifier with linear kernel is used in every case to train and make predictions with the new corpus.

\section{Experiments and results}
%\todo[author=HJ,inline]{Antes de describir los experimentos, describir los datos: cuantos corpora, de qu\'e tareas, cuantas features, etc. }

%\subsection{Data}
For the experimental evaluation we collected 81 publicly available text corpora, each associated with a different classification task, most of which can be categorized as one of 6 common NLP tasks: authorship attribution, author profiling, topic/thematic classification, irony and deception detection. Some of theses datasets are commonly used for benchmarks in text classification (e.g. Amazon, Dbpedia, 20NGs) while others have been used in competitions. %can be found in competition sites such as Kaggle and SemEval. 
%They also have diverse origins that include: social media, poems, news articles, songs, and TV shows. 
After processing each corpus to share the same format and codification we extracted the 72 meta-features for each of the 81 collections. To accelerate the meta-feature extraction process we limited the number of documents to 90,000 per category. The resultant matrix of size 81$\times$72 comprises our \textit{knowledge base} characterizing  multiple corpora.

%\subsection{Setup}
In an offline phase, for each classification task every representation was used for training and testing a classification model, the \textit{performance} of each representation was calculated with 3-fold Cross validation, they were also ranked from best (1) to worst (60).

\begin{figure}[htbp]
    \centering
    \includegraphics[scale=0.2]{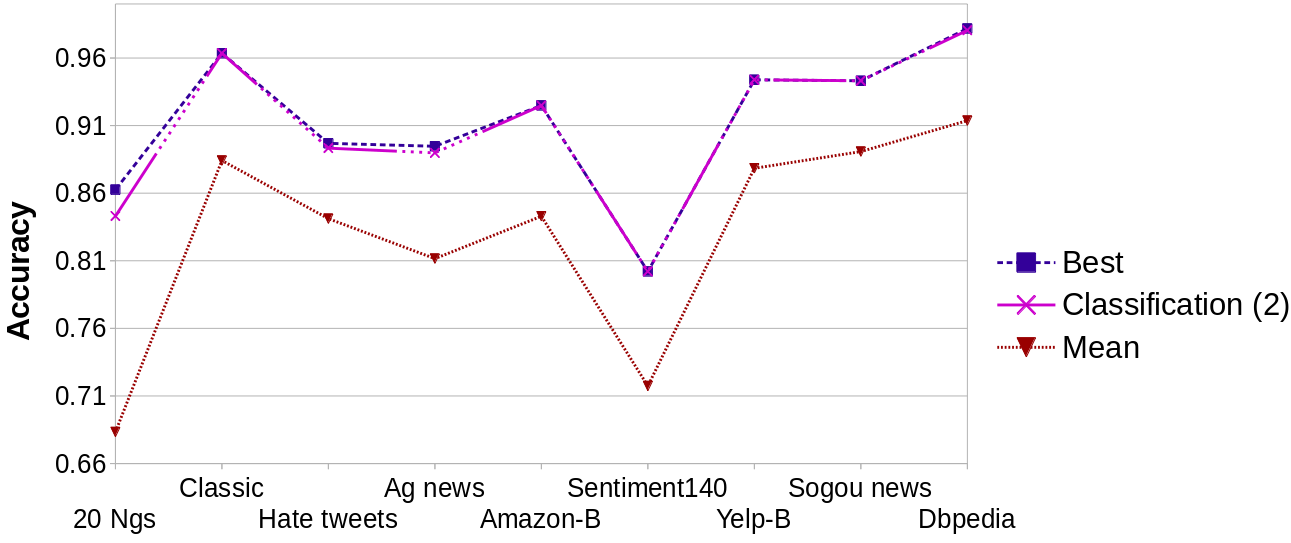}
   \caption{Accuracy of (2) in 9 selected corpora. }
    \label{fig:results}
\end{figure}

We evaluated the 4 meta-learning strategies with unseen tasks following a leave-one-out setting, using the results from 60 representations in the rest of the tasks as knowledge to decide which representation to recommend. The objective for the strategies, then, is to select what in exhaustive search was found to be the \textit{best} representation. We compared the average performance achieved by our strategies in 5 runs against the best solution found %obtained exhaustively 
and %against 
the  average performance of all of the considered  representations. Table \ref{tab:results} shows the average performance for each strategy after 5 runs in terms of the average accuracy and average rank. Figure \ref{fig:results} depicts the performance of our method and the baselines in 9 corpora (we selected these representative corpora to cover a wide variety of tasks and  because they are well known benchmarks). %frequently used for benchmark in text classification.

%\todo[author=HJ,inline]{por qu\'e est\'as 9 colecciones? decir/motivar en texto}

\begin{table}[htbp]
\caption{Average accuracy [0,1] and average rank [60,1] of different strategies in 81 corpus, the last row indicates the number of times the best representation was predicted. (1) Nearest corpus, (2) classification, (3) performance regression,  (4) rank prediction.}
\begin{center}\footnotesize{
\begin{tabular}{rrrrrrr}
\hline
Method & \multicolumn{1}{c}{Best} & \multicolumn{1}{c}{(1)} & \multicolumn{1}{c}{(2)} & \multicolumn{1}{c}{(3)} & \multicolumn{1}{c}{(4)} & \multicolumn{1}{c}{Random} \\ 
\hline
Avg Accu & 77.06$\pm$0 & 73.75$\pm$0 & \textbf{75.25$\pm$0.12} & 73.34$\pm$0.34 & \textbf{75.20$\pm$0.07} & 68.45$\pm$0 \\ 
Avg Rank & 1.00$\pm$0 & 14.20$\pm$0 & \textbf{8.71$\pm$0.46} & 14.30$\pm$1.31 & \textbf{8.51$\pm$0.34} & 30.30$\pm$0 \\ 
\# of 1s & 81.00$\pm$0 & 17.00$\pm$0 & \textbf{25.80$\pm$0.45} & 4.20$\pm$0.84 & 14.80$\pm$0.84 & 0.00$\pm$0 \\ 
\hline
\end{tabular}}
\end{center}
\label{tab:results}
\end{table}

The 4 strategies clearly outperform selecting a random representation and while in terms of average ranking they could be closer to the optimal, the average accuracy of (2) and (4) strategies was only 2\% behind the best. (2) also found the best representation 35\% of the time. Results show strong evidence that our meta-learning approach finds relations between corpora and pipeline performances that exploits prior knowledge for the autonomous classification of texts. %To further illustrate the differences of the strategies Figure \ref{fig:results3} compares their average ranking in our 9 selected corpora.

From the 72 proposed meta-features we tested different subsets according to their Gini importance from the Random Forest used in strategy (2). A subset of 38 meta-features improved our results relatively by 8\% with (2) and 38\% with (1) in terms of average ranking. We also compared this subset against a subset comprised of 19 \textit{traditional} meta-features used in related work. Using strategy (2) our subset outperformed the \textit{traditional} one by almost 0.8\% in average accuracy and 3 places in average rank. The results also showed a significant difference between using both subsets (p$<$.001 Student's t-test) The subset of 28 meta-features is detailed in Table \ref{tab:selection}.

%\begin{figure}[htbp]
%    \centering
%    \includegraphics[scale=0.4]{box.png}
%   \caption{Average average ranking of (2) with different subsets of meta-features: \textit{traditional}, all 72, and 38 selected. }
%    \label{fig:results3}
%\end{figure}

\begin{table}[htbp]
\caption{38 Meta-features selected by Gini importance}
\begin{center}
\begin{tabular}{l}
\hline
Meta-feature selection \\ 
\hline
average word length \\ 
document per category: \\ 
\quad min \\ 
\quad max \\ 
\quad average \\ 
\quad standard deviation \\ 
\quad average / stdev \\ 
\quad entropy \\ 
word per document: \\ 
\quad average \\ 
\quad skewness \\
\quad entropy \\
Imalance Degree \\ 
SEM \\ 
UVB \\ 
SVB \\ 
MRH\_J \\ 
VDR \\
max vocabulary \\ 
average vocabulary \\ 
sd vocabulary \\ 
skweness vocabulary \\ 
avg/stdev vocabulary \\ 
pca: \\ 
\quad singular values sum \\ 
\quad explained ratio \\ 
\quad explained variance \\ 
\quad explained variance (1) \\ 
\quad pca max \\ 
\quad pca skewness \\ 
\quad pca kurtosis \\ 
data sparsity \\
data separability \\ 
linear separability \\ 
\% of zeros \\ 
\% of adpositions \\ 
\% of adverbs \\ 
\% of conjunctions \\ 
\% of nouns \\ 
\% of numbers \\ 
\% of untagged words\\
difficult words\\
\hline
\end{tabular}
\end{center}
\label{tab:selection}
\end{table}

%\begin{figure}[htbp]
%   \centering
%    \includegraphics[scale=0.26]{barra.png}
%   \caption{\small{Comparison of 4 proposed %strategies in 9 corpora.}}
%    \label{fig:results3}
%\end{figure}

\begin{figure}[htbp]
    \centering
    \includegraphics[scale=0.2]{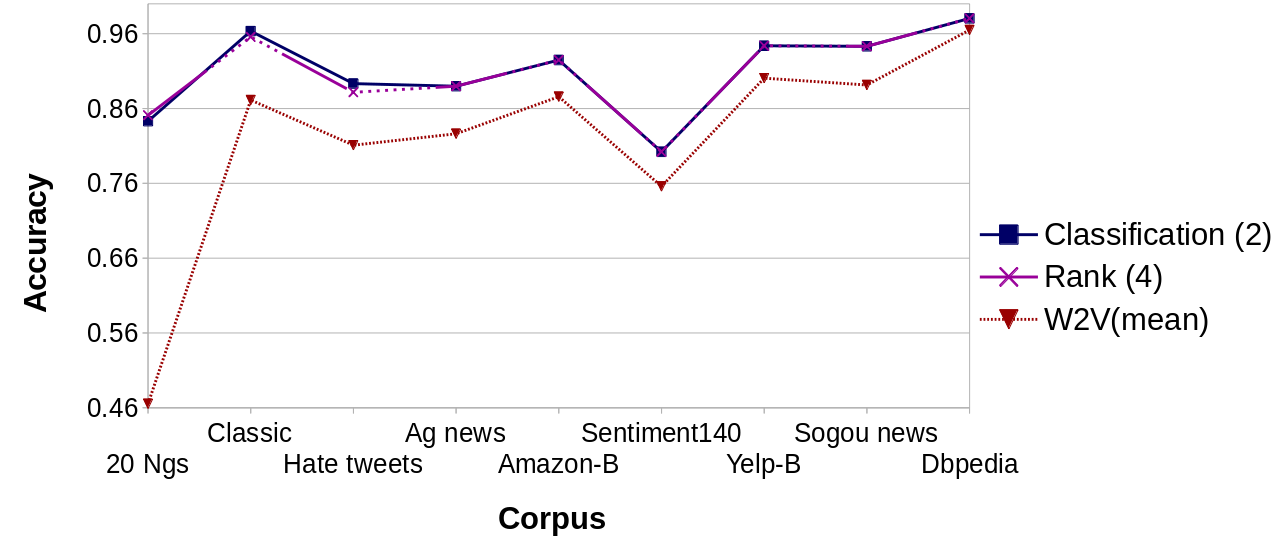}
   \caption{Accuracy comparison between (2), (4) and Word2Vec in 9 corpora. }
    \label{fig:results2}
\end{figure}

In addition, we compared our strategies with commonly used representations such as pre-trained Word2Vec and Bag-of-Words outperforming them in average by 9\% and 3\% respectively, Figure \ref{fig:results2} depicts this comparison (between strategy (4) and W2V) in the 9 corpora we selected. Despite the robustness of such common representations their performance can usually be improved by fine tuning some of their hyper-parameters or they are largely outperformed by another, as shown in the results the strategies are able to find these improvements.

%\begin{figure}[htbp]
%    \centering
%    \includegraphics[scale=0.26]{chart8.png}
%   \caption{\small{Accuracy comparison between (2), (4) and Word2Vec in 9 corpora.}}
%    \label{fig:results2}
%\end{figure}

\section{Conclusion and future work}
We introduced a meta-learning method that takes as input a corpus and without human intervention builds a model to solve a text classification task focusing on the selection of a vector-based  representation. % and based mainly on meta-learning techniques. 
The results show empirically that  this approach is   able to characterize tasks and approximate an optimal representation. Our work can be used to \textit{warm-start} an optimization technique with our proposed meta-learning setting allowing us to expand the search space and ideally finding pipelines that perform better than those designed by humans. Our also work comprises a first step towards the automated recommendation of full text classification pipelines. The source code of our method is available under an open source license at: https://github.com/jorgegus/autotext. %, two approaches could be explored in this direction: combining the pre-processing methods of NLP with the models and hyperparameters from machine learning in the same search space or optimizing a classification model after the representation vector is selected. 

% Acknowledgements should go at the end, before references and appendices

\bibliographystyle{unsrt}  
\bibliography{references}  %%% Remove comment to use the external .bib file (using bibtex).


\end{document}